\documentclass[letterpaper, 10 pt, conference]{ieeeconf}  

\IEEEoverridecommandlockouts                              

\usepackage{amsmath,amsfonts,amssymb}
\usepackage{algorithm}
\usepackage{algorithmic}

\usepackage{array}
\usepackage[caption=false,font=normalsize,labelfont=sf,textfont=sf]{subfig}
\usepackage{textcomp}
\usepackage{stfloats}
\usepackage{url}
\usepackage{verbatim}
\usepackage{graphicx}
\usepackage{cite}

\usepackage{xcolor}
\usepackage{makecell}
\usepackage{marvosym}
\usepackage{float}
\usepackage{hyperref}
\def\degree{${}^{\circ}$}
\def\BibTeX{\text{B\kern-.05em{\sc i\kern-.025em b}\kern-.08em
    T\kern-.1667em\lower.7ex\hbox{E}\kern-.125emX}}
\newcommand{\instfont}{\fontfamily{cmss}\selectfont}
\newcommand{\thhead}[1]{\textbf{#1}}
\newcommand{\thunit}[1]{\fontsize{7.0pt}{\baselineskip}\selectfont\textbf{\sffamily #1}}
\newcounter{magicrownumbers}
\newcommand\rownumber{\stepcounter{magicrownumbers}\arabic{magicrownumbers}}
\usepackage{booktabs}
\usepackage{multirow}
\usepackage{colortbl}
\usepackage{fbox}
\definecolor{Gray}{gray}{0.85}
\newcolumntype{g}{m{0.038\textwidth}<{\centering}}
\newcolumntype{z}{>{\columncolor{lightgray}}Y}
\usepackage{geometry}
\title{\LARGE \bf
MLANet: Multi-Level Attention Network with Sub-instruction for Continuous Vision-and-Language Navigation
}

\author{Zongtao He, Liuyi Wang, Shu Li, Qingqing Yan, Chengju Liu and Qijun Chen, \IEEEmembership{Senior Member, IEEE}
\thanks{This work was supported by the National Natural Science Foundation of China under Grants 61733013, 62073245, 62173248; Suzhou Key Industry Technological Innovation-Core Technology R\&D Program, No. SGC2021035}
\thanks{The authors are with the Robotics and Artificial Intelligence Lab (RAIL), Tongji University, Shanghai, 201804, China. \protect E-mail:\{xingchen327, wly, lishu, qyan\_0131, liuchengju, qjchen\}@tongji.edu.cn.}%
}
\begin{document}

\maketitle
\thispagestyle{empty}
\pagestyle{empty}

\begin{abstract}
Vision-and-Language Navigation (VLN) aims to develop intelligent agents to navigate in unseen environments only through language and vision supervision.
In the recently proposed continuous settings (continuous VLN), the agent must act in a free 3D space and faces tougher challenges like real-time execution, complex instruction understanding, and long action sequence prediction.
For a better performance in continuous VLN, we design a multi-level instruction understanding procedure and propose a novel model, Multi-Level Attention Network (MLANet).
The first step of MLANet is to generate sub-instructions efficiently.
We design a Fast Sub-instruction Algorithm (FSA) to segment the raw instruction into sub-instructions and generate a new sub-instruction dataset named ``FSASub".
FSA is annotation-free and faster than the current method by 70 times, thus fitting the real-time requirement in continuous VLN.
To solve the complex instruction understanding problem, MLANet needs a global perception of the instruction and observations.
We propose a Multi-Level Attention (MLA) module to fuse vision, low-level semantics, and high-level semantics, which produce features containing a dynamic and global comprehension of the task.
MLA also mitigates the adverse effects of noise words, thus ensuring a robust understanding of the instruction.
To correctly predict actions in long trajectories, MLANet needs to focus on what sub-instruction is being executed every step.
We propose a Peak Attention Loss (PAL) to improve the flexible and adaptive selection of the current sub-instruction.
PAL benefits the navigation agent by concentrating its attention on the local information, thus helping the agent predict the most appropriate actions.
We train and test MLANet in the standard benchmark.
Experiment results show MLANet outperforms baselines by a significant margin. \footnote{Codes and dataset are available at \url{https://github.com/RavenKiller/MLA}}
\end{abstract}


\section{Introduction}
\label{sec:introduction}
Service robots with the ability to follow natural language instructions have been a longstanding subject of robotic research. 
Towards this goal, Vision-and-Language Navigation (VLN) \cite{anderson8578485} aims at developing an embodied agent with the ability to navigate in an unstructured environment through linguistic supervision and visual feedback.
Current VLN research is mainly conducted on simulated environments, which can be divided into two types: discrete and continuous.
In discrete settings such as R2R \cite{anderson8578485}, REVERIE \cite{qi9156641} and RxR \cite{ku2020room}, the agent is only allowed to reach pre-defined graph nodes and executes ideal teleportation between nodes.
While in continuous setting \cite{krantz2020beyond}, the agent can move freely on a continuous space and executes speed control actions at every step.
Continuous VLN is a valuable research direction because it simulates robots' behavior in the real world and does not rely on graph assumptions.
However, continuous VLN faces challenges like real-time execution, complex instruction understanding, and long action sequence prediction, thus requiring a more robust capacity to exploit the instruction.

\begin{figure}[t]
\centering
\includegraphics[width=0.48\textwidth]{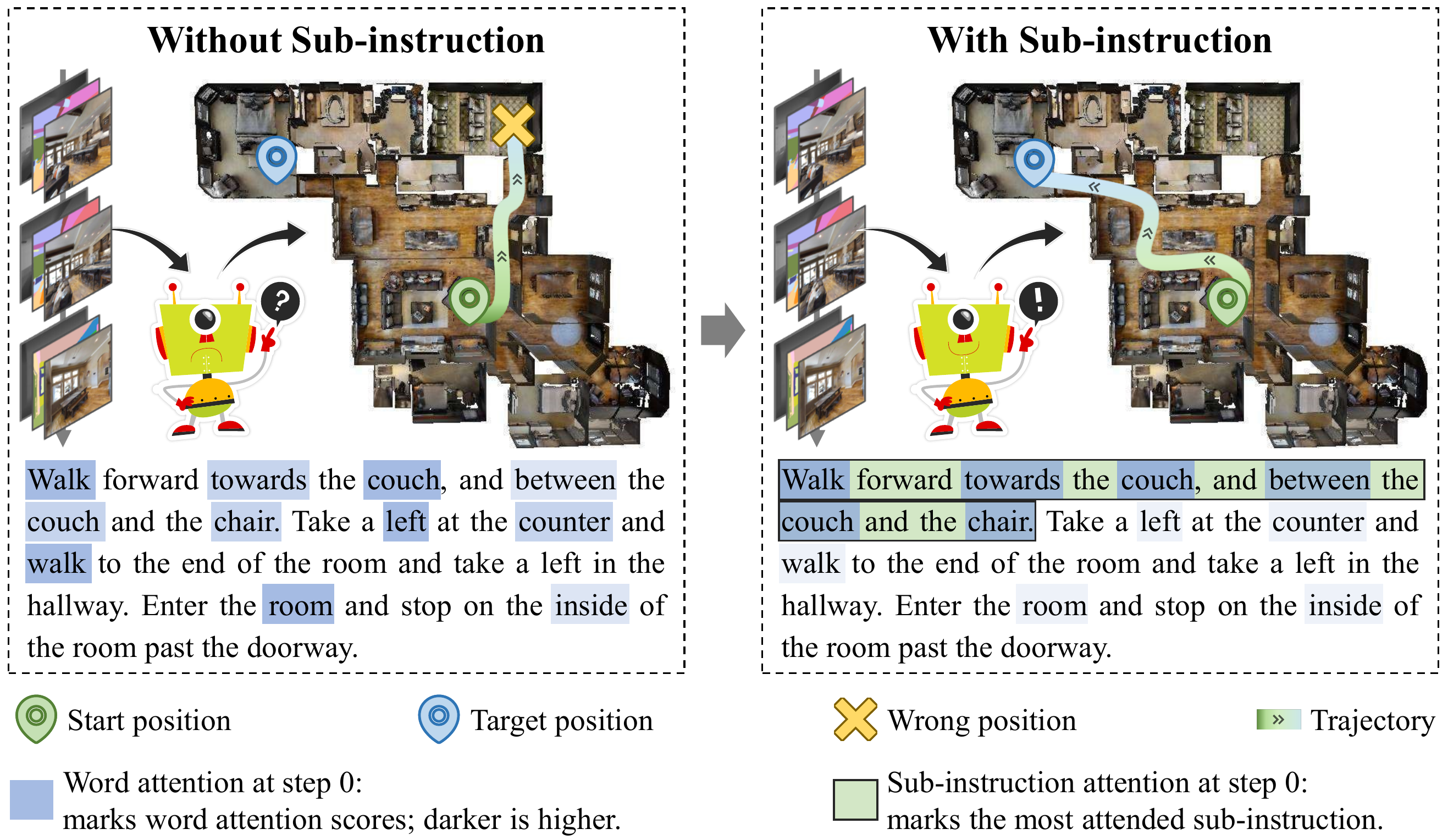}
\caption{Improvement by leveraging sub-instructions. With only word attention (left), the agent misunderstands the task at initial steps and finally reaches a wrong position. Sub-instruction (right) strengthens useful information and weakens the impact of noise words (``room" in the third sentence is noise when conducting initial steps), so the agent chooses a correct direction and successfully navigates to the target position.}
\label{fig:vln_intro}
\end{figure}
Multi-level instruction understanding with sub-instruction is an inspiring solution for challenges in continuous VLN.
Imagining a human required to finish a complex navigation instruction, what would he/she do?
It is natural that he/she will segment the instruction into several sub-instructions and complete them one by one.
The significance of sub-instruction is that it brings navigation information from a different semantic granularity and breaks a complex task into easy-to-finish ones.
In this paper, we use ``level" to represent the semantic granularity of a language unit.
Every single word only contains very limited information, which we call low-level semantics, while sub-instruction brings complete information for each part of the navigation path, which we called high-level semantics.
Previous VLN methods \cite{krantz2020beyond,wang2018look,Hong2020bentity,chi2020just,xia2020multi,fried2018speaker,tan2019learning} mainly use specific encoders to extract word features which contain relatively low-level semantics, neither treating a sub-instruction as a whole nor leveraging the high-level semantics in sub-instructions.
Such neglect may bring a word grouping problem, which means that the agent cannot correctly group meaningful words together.
For example, in the left part of Fig. \ref{fig:vln_intro}, the agent pays attention to some noise words (``left" in the second sentence; ``room" in the third sentence), so it is hard for the agent to figure out what to do.
Although some works \cite{hong2020sub,zhang2021spc-nav,zhu-etal-2020-babywalk} make initial efforts to utilize sub-instructions, they still stay at extracting word features in a sub-instruction rather than high-level features.
Furthermore, these methods do not go further into continuous VLN to unleash sub-instructions' potential.
If we simultaneously leverage low-level and high-level semantics to form a multi-level instruction understanding, like the right part in  Fig. \ref{fig:vln_intro}, the injection of sub-instruction information can relatively mitigate noise words and gives more precise guidance for the action prediction.

To improve the performance in continuous VLN, we design a multi-level instruction understanding procedure and propose a novel Multi-Level Attention Network (MLANet).
We first design an efficient and annotation-free algorithm, Fast Sub-instruction Algorithm (FSA), to segment the raw instruction into sub-instructions.
Previous sub-instruction algorithms \cite{hong2020sub,zhang2021spc-nav} contain heavy components like deep neural networks, which suffer from calculation efficiency problems and are impossible for real-time applications.
In contrast, FSA is highly efficient because of our streamlined design, which can be used for both dataset pre-processing and real-time navigation.
We also release a dataset, ``FSASub", processed by FSA publicly for future research.
Then, the Multi-Level Attention (MLA) module is used to co-ground multi-modal features and fuse multi-level instruction features.
On the one hand, MLA supplies precise cognition of the navigation process to the agent and helps the agent conduct reasonable actions.
On the other hand, the fusion of multi-modal and multi-level features forms a comprehensive perception of the instruction, giving the agent a richer context of the navigation path.
Instead of only using word features in one sub-instruction \cite{hong2020sub,zhang2021spc-nav}, MLA actually utilizes sub-instruction features and supplies a global understanding of the complex instruction. 
Finally, we propose an auxiliary loss function named Peak Attention Loss (PAL) for the adaptive sub-instruction selection.
In the navigation process, the agent may consider wrong or multiple sub-instructions, which is different from our expectations.
PAL shapes the attention score to a single ``peak" pattern and helps the agent to build a local perception of the current sub-instruction so that the agent can predict every action without distractions in long trajectories.
Equipped with FSA, MLA, and PAL, MLANet mitigates the time cost of the sub-instruction generation and obtains both global and local perceptions of the instruction, overcoming challenges like complex instruction understanding and long action sequence prediction.

In summary, our contributions are as follows:
\begin{itemize}
\item We propose an efficient FSA for the generation of sub-instructions, which is annotation-free and faster than the previous method by 70 times.
\item We propose MLA for comprehensive multi-modal co-grounding and multi-level semantics fusion, which effectively align the instruction and observations and benefit the robust global understanding of the complex instruction.
\item We design PAL for a flexible and adaptive sub-instruction selection, which benefits the local perception of the current sub-instruction and helps the agent produce reasonable actions in long trajectories.
\item Equipped with FSA, MLA, and PAL, we construct a novel model, MLANet. Experiments on the VLN-CE benchmark show that MLANet outperforms baselines by a significant margin. We release the code together with the FSASub dataset for future research.
\end{itemize}

\section{Related work}
\label{sec:related}
\subsection{VLN in Continuous Environments}
\label{sec:related:vlnce}
The ultimate purpose of VLN is the embodied navigation agent and real-world application.
To achieve this goal, several works\cite{irshad2021hierarchical,krantz2020beyond,kolve2017ai2} try to raise unrealistic assumptions in R2R-based datasets and put the VLN task into continuous environments.
VLN-CE \cite{krantz2020beyond} is the first attempt at building a more realistic setting for VLN. 
In VLN-CE, the agent perceives the environment by a single, forward-mounted RGBD camera, which differs from panoramic observation in most R2R-based works \cite{fried2018speaker,ma2019self,tan2019learning,majumdar2020improving,ZHAN202368}. 
Due to the continuous state space, long navigation steps and the limited egocentric view, the task becomes more challenging.
Robo-VLN \cite{irshad2021hierarchical} tries to solve more complicated VLN task with longer trajectory lengths and challenging obstacles by hierarchical reinforcement learning.
Li et al. \cite{LI2021368} leverages meta-learning to improve the generalization on multi goals and solves visual mapless navigation in AI2-THOR\cite{kolve2017ai2}.
As for VLN-CE, there are several methods \cite{Georgakis_2022_CVPR,raychaudhuri2021language,krantz2021waypoint} proposed to solve the continuous VLN task.
However, they do not pay much attention to instruction understanding and only leverage word features.
Our work proposes introducing sub-instruction to form a multi-level instruction understanding in continuous VLN.
Another improvement we make is employing CLIP \cite{radford2021learning} as the image encoder, which has more generalization than ResNet \cite{he2016deep} adopted in previous methods.

\subsection{Attention Mechanism in VLN}
\label{sec:related:attention}
There exist two ways to utilize the attention mechanism in VLN.
The first is using soft attention as a cross-modal unit.
For example, Speaker-Follower \cite{fried2018speaker} and  EnvDrop \cite{tan2019learning} leverage one or more soft attention units to co-ground vision and instruction information.
The second way uses pre-trained multi-modal Transformer \cite{vaswani2017attention,HAN202389}, which is made of several self-attention layers.
VLNBERT \cite{majumdar2020improving} uses cross-modal BERT to select the correct navigation path.
Recurrent VLNBERT \cite{hong2021vln} introduces the recurrent mechanism for the action decoding process.
However, existing methods always apply attention mechanisms to word features and neglect sub-instruction features.
In this paper, we propose a MLA to pay more attention to the fusion of multi-modal and multi-level features.
Compared with previous works, we utilize multi-head attention as the basic unit of MLA instead of simple soft attention, strengthening the ability to align information from different modalities.
Another different point is that MLA actually leverages sub-instruction features to supply more robust guidance for the agent and improves the global perception of the instruction.
Therefore, MLA is constructed by existing units but develops a new idea for utilizing the navigation instruction.

\subsection{Sub-instruction}
\label{sec:related:subinst}
The existing VLN methods seldom consider the structural characteristics of language instructions.
Classical methods \cite{anderson8578485,fried2018speaker} mainly use LSTM to extract word features from the raw instruction and then apply attention to the subsequent process.
Recent methods \cite{li2019robust,majumdar2020improving} introduce Transformer to process language but still only consider raw instructions.
FGR2R \cite{hong2020sub} is the first effort to understand sub-instructions, which only takes one sub-instruction every step and also decides whether to switch sub-instruction.
However, the hard switching strategy lacks global perception of the instruction and cannot recover from failure steps.
BabyWalk \cite{zhu-etal-2020-babywalk} records pasted sub-instructions in the history buffer to form a global perception.
SpC-NAV \cite{zhang2021spc-nav} proposed to use indicator words to distinguish and represent sub-instructions.
However, previous methods \cite{hong2020sub,zhu-etal-2020-babywalk,zhang2021spc-nav} all stay at processing word features and do not form a multi-level instruction understanding.
Moreover, they are designed for discrete environments and ignore the time efficiency of the sub-instruction generation, which is an essential factor in real-world applications. 

This paper proposes a complete procedure to utilize sub-instructions.
First, we design an efficient algorithm FSA to segment the raw instruction into sub-instructions, which do not rely on expert annotations and can process any new instruction in real time.
Instead of the switch strategy, we leverage sub-instruction features and use multi-head attention to fuse sub-instruction information dynamically, ensuring a global perception of the instruction.
Finally, we propose PAL to shape the attention score to promote the model's concentration on the current sub-instruction, benefitting the local understanding of the instruction.

\section{MLANet}
\label{sec:method}

\begin{figure*}[t]
\centering
\includegraphics[width=0.96\textwidth]{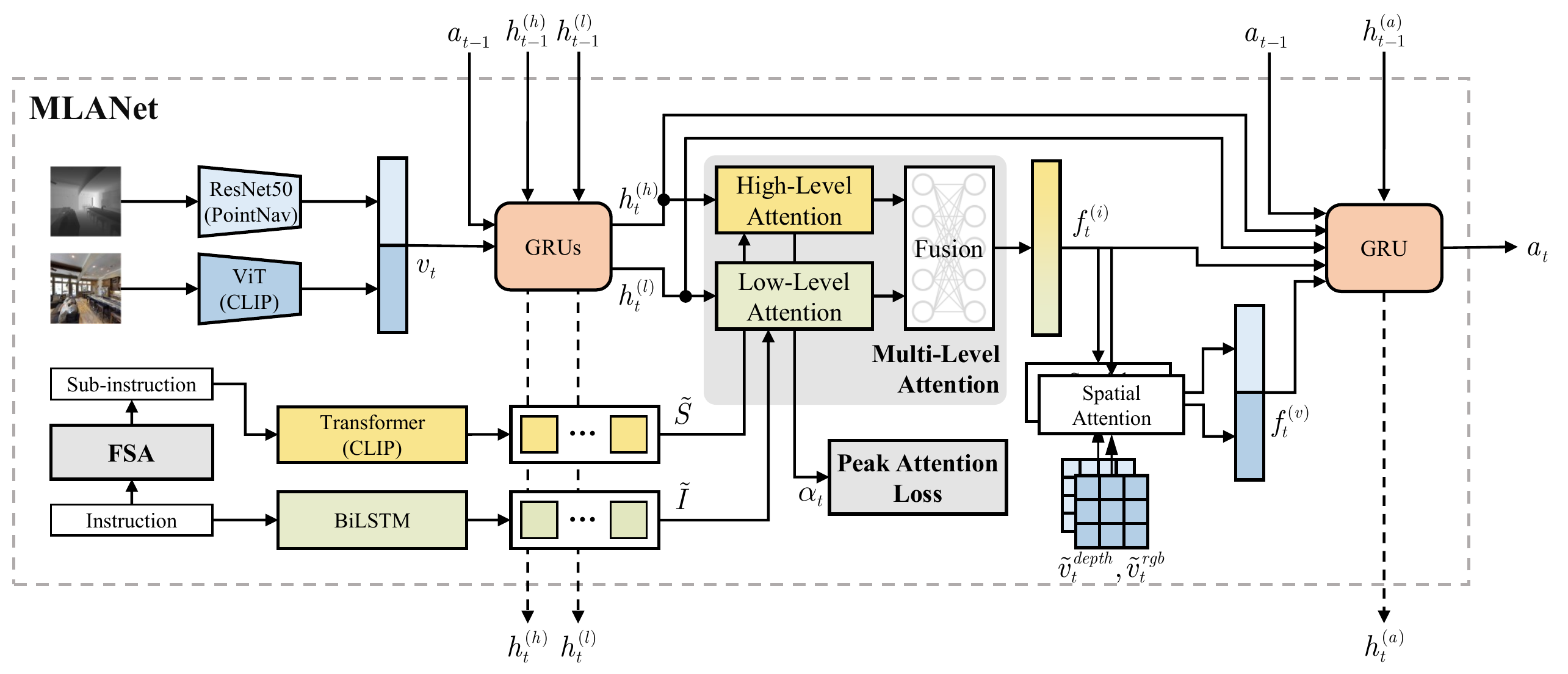}
\caption{An illustration of the model architecture. The input contains an RGB image, a depth image, a raw instruction, and sub-instructions from FSA. After encoders and GRU memory, instruction features are fused by MLA, and vision features are fused by spatial attention. The action decoder accepts all hidden features and outputs action. Attention score of high-level attention will be used for PAL. Gray shading and bold fonts mark important components.}
\label{fig:model_architecture}
\end{figure*}
This section gives our proposed model MLANet.
As shown in Fig. \ref{fig:model_architecture}, MLANet contains following components:
\begin{itemize}
    \item[--] \textbf{Fast Sub-instruction Algorithm} that supplies sub-instruction efficiently for the navigation agent.
    \item[--] \textbf{Vision and Language Encoders} that embed raw inputs from different modalities into dense feature spaces.
    \item[--] \textbf{Multi-Level Attention Module} that fuse multi-modal features and multi-level instruction features to form a global understanding of the instruction.
    \item[--] \textbf{Peak Attention Loss} that helps shape the high-level attention distribution and promotes the local perception of the current navigation progress.
    \item[--] \textbf{Navigation Reasoning Decoder} that accepts all latent features and outputs the predicted action.
\end{itemize}
In this section, we will state the problem formulation, outline the model architecture, and then focus on several important components in MLANet.
\subsection{Problem Formulation}
\label{sec:method:formulation}
In VLN, the agent's goal is navigating to the target position through language supervision and visual feedback.
During the navigation process, the agent receives a consistent instruction $I=\{i_1, \dots, i_L\}$, where $i_l$ denotes the $l$-th word and $L$ is the length of the instruction.
In our method, the agent also receives a sub-instruction sequence $S=\{s_1, \dots, s_N\}$, where $s_n$ denotes $n$-th sub-instruction and $N$ is the number of sub-instructions.
Raw instructions are segmented into sub-instructions by FSA, as described in Algorithm \ref{alg:sub_cut}.
Our FSA is feasible to be applied in any navigation model, with the characteristics of plug and play.

At every time step $t$, the agent perceives a 90\degree, single-view RGB image $o^{rgb}_t\in \mathbb{R}^{256\times 256\times 3}$ and depth image $o^{depth}_t\in \mathbb{R}^{256\times 256}$.
The agent uses an action set containing four actions: $\{\text{move forward 0.25m}, \text{turn left 15\degree}, \text{turn right 15\degree}, \text{stop}\}$.
Let $\hat{a}_t$ be the ground-truth action and $a_t$ be the predicted action.
After executing the action $a_t$, the position of the agent changed from $p_{t-1}$ to $p_t$, where $p_t\in \mathbb{R}^3$ represents position vector in the continuous environment.

For a given start position $P_s$ and end position $P_e$, the path can be judged as finished when the final position of the agent $p_T$ satisfies $\left\|P_e-p_T\right\|\leq \epsilon$, where $T$ is the time the agent predict to stop and $\epsilon=\text{3m}$ is the threshold distance \cite{anderson8578485}.
In summary, a VLN task can be abstracted as a sequence-to-sequence problem: the agent is given sequence inputs $I, S, O^{rgb}, O^{depth}$ and is expected to output an action sequence $A=\{a_t\}^T_{t=1}$, which allows the agent moving from the start position $P_s$ to the end position $P_e$. 

\subsection{Model Architecture Overview}
\label{sec:method:formulation}
Here we describe the overall architecture of MLANet.
As shown in Fig. \ref{fig:model_architecture}, MLANet uses the encoder-decoder framework.

\textbf{The encoding part}. Vision and language encoders are used to process raw inputs into dense features, which can utilize models pretrained on large-scale datasets.
The inputs of MLANets are depth image $O^{depth}$, RGB image $O^{rgb}$, instruction $I$, and sub-instructions $S$ from FSA (detailed in Sec \ref{sec:method:ssa}).
So, there are four encoders to process different modalities: depth encoder, RGB encoder, instruction encoder, and sub-instruction encoder, which are located at the left part in Fig. \ref{fig:model_architecture}.

The environmental observations are depth images and RGB images.
For the depth image $O^{rgb}$, we employ ResNet \cite{wijmans2019dd} as the depth encoder to extract the spatial depth features $\tilde{v}^{depth}_t$ and use a fully-connected network to produce a low-dimension and learn-able feature $v^{depth}_t$
For the RGB image, we propose to use a ViT from CLIP \cite{radford2021learning} as the RGB encoder to extract the  spatial RGB features $\tilde{v}^{rgb}_t$.
After ViT mapping, features at the [CLS] position contain compressed visual information.
So we further pass [CLS] features to a fully-connected network to produce a low-dimension feature $v^{rgb}_t$.
History information is essential for navigation, and we use memory units to remember environmental observations, which will be detailed in \ref{sec:method:mla}.

The language modality contains two levels of semantic units: words and sub-instructions, and we use different embedding strategies for them.
For the raw instruction containing words $I$, we use a BiLSTM as the instruction encoder to obtain low-level instruction features $\tilde{I}$ and keep the tracing of word meanings.
For sub-instructions, we employ the text Transformer from CLIP \cite{radford2021learning} as the sub-instruction encoder to obtain high-level instruction features $\tilde{S}$.
$\tilde{I}$ and $\tilde{S}$ are then projected to low-dimensional spaces for learnable features.

\textbf{The attention part}. After encoders, how to effectively aggregate information from different modalities is a key problem.
We propose MLA to dynamically fuse vision states $h^{(h)}_t,h^{(l)}_t$ and multi-level instruction features $\tilde{I},\tilde{S}$ to produce fused feature $f^{i}_t$ (detailed in Sec \ref{sec:method:mla}).
And we propose PAL to supervise the high-level score $\alpha_t$ in MLA for a better local perception of the instruction (detailed in Sec \ref{sec:method:pal}).
Following VLN-CE \cite{krantz2020beyond}, we further use spatial attention modules to generate a fused visual feature $f^{v}_t$ from spatial visual features.

\textbf{The decoding part}. Finally, all latent features are inputted to a recurrent action decoder.
The action decoder is a GRU module and produces an action state $h^{(a)}_t$, which will be used to generate the predict action $a_t$.

\subsection{Fast Sub-instruction Algorithm}
\label{sec:method:ssa}

\begin{algorithm}[tb]
\caption{Fast Sub-instruction Algorithm}  
\label{alg:sub_cut}  
\begin{algorithmic}[1]
\REQUIRE Instruction set $D=\{I_i\}_{i=1}^N$
\ENSURE Sub-instruction set $O=\{(S_i,\ \texttt{tokens}_i)\}_{i=1}^N$
\STATE Initialize $O \gets \{\}$
\FOR{$I$ in $D$}
\STATE Filter out invalid characters in $I$, such as $\backslash$r$\backslash$n;
\STATE Filter out invalid phrase in $I$, such as HTML marks;
\STATE $\texttt{subs} \gets$ Segment $I$ by coarse sentence tokenizer;
\STATE $S \gets \{\}$;
    \FOR{$\texttt{sub}$ in $\texttt{subs}$}
    \STATE $\texttt{words},\ \texttt{tags} \gets \text{POSTagger}(\texttt{sub})$;
    \STATE $\texttt{sub\_now} \gets \emptyset$;
    \STATE $N \gets$ The size of \texttt{words};
    \FOR{$i \gets 1$ to $N$} 
            \IF{NeedRefine(i, $\texttt{tags}, \texttt{words})$}
                \STATE Add \texttt{sub\_now} into $S$;
                \IF{$\texttt{wrods}[i]$ is not punctuation}
                    \STATE $\texttt{sub\_now} \gets \texttt{words}[i]$;
                \ELSE
                    \STATE $\texttt{sub\_now} \gets \emptyset$;
                \ENDIF
            \ELSE
                \STATE $\texttt{sub\_now} \gets \texttt{sub\_now}+\texttt{words}[i]$;
            \ENDIF
        \ENDFOR
        \IF{\texttt{sub\_now} is not $\emptyset$}
            \STATE Add \texttt{sub\_now} into $S$;
        \ENDIF
    \ENDFOR
    \STATE $\texttt{tokens} \gets$ Tokenize words for every element in $S$;
    \STATE Add ($S$,\ \texttt{tokens}) into $O$;
\ENDFOR
\end{algorithmic}
\end{algorithm}

Raw instructions consist of several words.
When applying the attention mechanism to word features, there may be word grouping or attention dispersion problems.
Sub-instructions benefit the navigation process because they supply high-level guidance and easy-to-finish goals.
Introducing sub-instructions is hopeful of mitigating the effect of noise words and overcoming word grouping problems.
The current sub-instruction chunking method \cite{hong2020sub} contains heavy components and suffers from long-time cost, which cannot fit real-time applications.
Guided by two principles: efficiency and simplicity, we design a novel FSA to segment the raw instruction into sub-instructions.
A simplified pseudo-code of FSA is shown in Algorithm \ref{alg:sub_cut}.
Firstly, the raw instruction is cleaned by several pre-defined rules, such as deleting escape characters.
Then, the cleaned instruction is segmented coarsely by a customized sentence tokenizer, which finds sentence boundaries and decides where to segment a long sentence.
However, due to the complexity of the navigation instruction, the simple sentence tokenizer may fail when encountering several conjunctions or ambiguous expressions.
So, we design a concise set of refining rules to refine sub-instructions according to part-of-speech tagging results.
This rule set contains several heuristic rules for judging where to start a new sub-instruction more precisely.
Finally, we tokenize every word according to the vocabulary, obtaining a sub-instruction set named ``FSASub".
FSA is a plug-and-play component, which means that FSA can also be embedded in the navigation model besides pre-processing the dataset.
Fig. \ref{fig:mla_detail} gives a qualitative example of how FSA works.

\subsection{Multi-Level Attention Module}
\label{sec:method:mla}
\begin{figure}[tbh]
\centering
\includegraphics[width=0.48\textwidth]{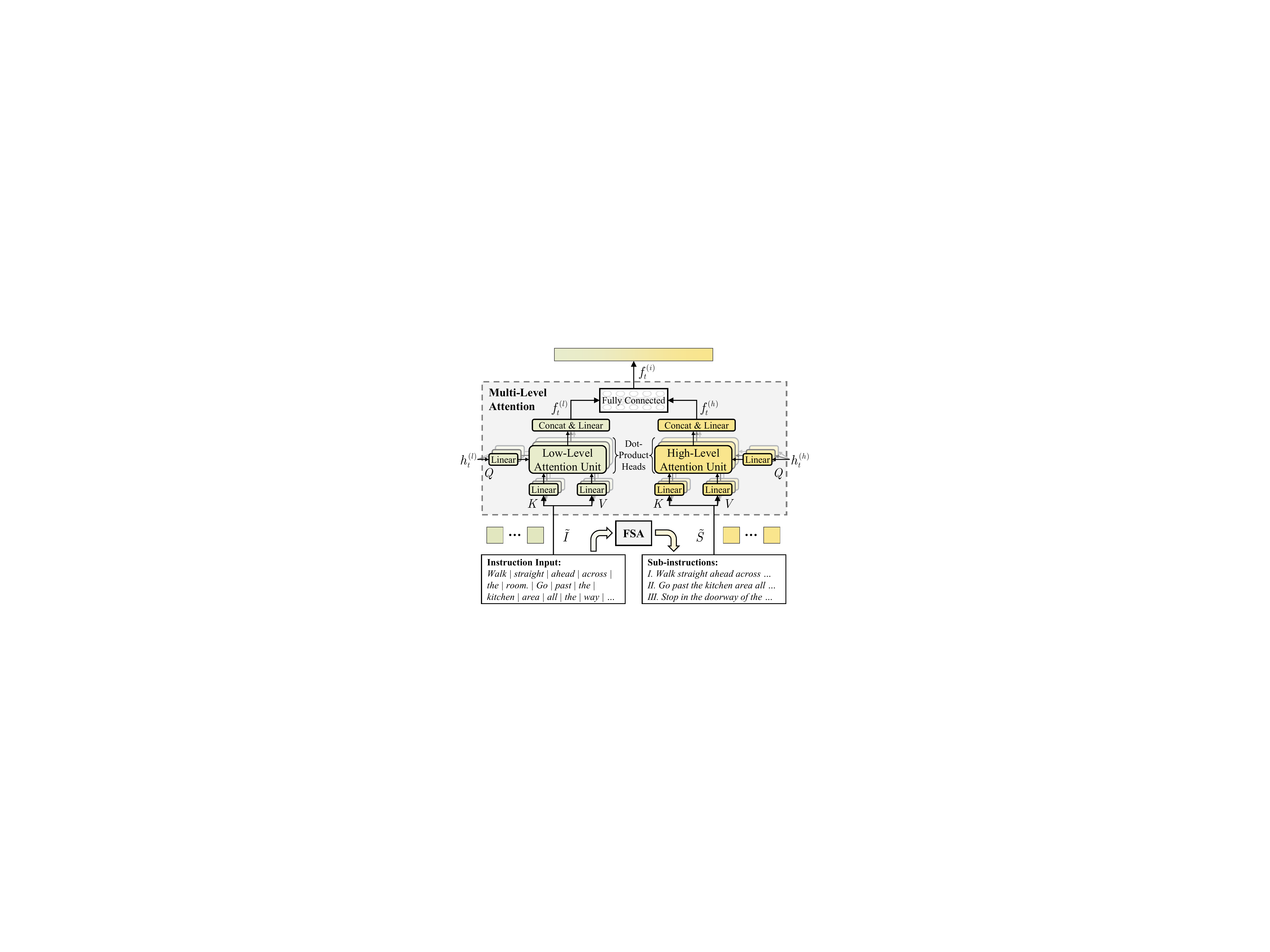}
\caption{An inner sight of the MLA module. Two visual hidden states are inputted as attention queries, and multi-level instruction features are inputted as keys and values. After attention blocks, high-level and low-level outputs are fused by a fully-connected layer to produce a dynamic global perception of the instruction.}
\label{fig:mla_detail}
\end{figure}

The critical function of MLA module is to fuse multi-modal and multi-level features, as shown in Fig. \ref{fig:mla_detail}.
Multi-modal features refer to features from different modalities, i.e., vision and language modalities.
Multi-level features refer to instruction features with different semantic levels.
The significance of MLA is that it considers not only low-level semantics $\tilde{I}$ from words but also leverages high-level features $\tilde{S}$ from sub-instructions.
Such multi-level perception of instructions can effectively alleviate unfavorable effects caused by dispersed word attention and forms a better alignment between vision states,  instructions, and agent actions.
After MLA module, the fused feature $f^{(i)}_t$ possesses a global perception to a certain extent at the temporal, spatial, and semantic levels.

In detail, the inputs of MLA are vision memory states $h^{(h)}_t,h^{(l)}_t$, low-level instruction features $\tilde{I}$ and high-level instruction features $\tilde{S}$.
$h^{(h)}_t,h^{(l)}_t$ come from two GRU units as:
\begin{equation}
\label{eq:encoder_gruh}
    h^{(h)}_t = \text{GRU}_h(\text{Concat}(v_t,\tilde{a}_{t-1});h^{(h)}_{t-1})
\end{equation}
\begin{equation}
\label{eq:encoder_grul}
    h^{(l)}_t = \text{GRU}_l(\text{Concat}(v_t,\tilde{a}_{t-1});h^{(l)}_{t-1}),
\end{equation}
The reason for using two memory units is that different semantic levels focus on different aspects.
Low-level vision state $h^{(l)}_t$ will tend to store objects or directions information corresponding to single words.
While high-level vision state $h^{(h)}_t$ will consider long-term trajectory information from sub-instructions.
Then, we use multi-head attention instead of soft attention to build attention blocks at different semantic levels.
Low-level instruction features $\tilde{I}$ and vision state $h^{(l)}_t$ are inputted to low-level attention part, with  $h^{(l)}_t$ as query and $\tilde{I}$ as key and value.
High-level instruction features $\tilde{S}$ and vision state $h^{(h)}_t$ are inputted to high-level attention part, with  $h^{(l)}_t$ as query and $\tilde{I}$ as key and value.
Two output features are concatenated and passed to a fully-connected network to aggregate semantics from different levels:
\begin{equation}
\label{eq:mla_h}
    f^{(h)}_t = \text{MultiHead}(h^{(h)}_t, \tilde{S}, \tilde{S})
\end{equation}
\begin{equation}
\label{eq:mla_l}
    f^{(l)}_t = \text{MultiHead}(h^{(l)}_t, \tilde{I}, \tilde{I})
\end{equation}
\begin{equation}
\label{eq:mla_cat}
    f^{(i)}_t = \text{FC}([f^{(h)}_t; f^{(l)}_t])
\end{equation}
where $\text{MultiHead}(Q,K,V)$ is the standard multi-head attention block \cite{vaswani2017attention}, $\text{FC}$ is a fully-connected layer and $f^{(i)}_t\in\mathbb{R}^{512}$ is the fused instruction feature.

What information does $f^{(i)}_t$ contains?
The MLA's multi-level structure helps inject precise and complete navigation process information, which serves as high-level guidance for corrective actions.
The fusion layer in MLA dynamically fuses multi-level features, relatively strengthening the current goal and weakening unrelated information.
In addition, the utilization of the multi-head attention mechanism further enriches the multi-level understanding.
With these designs, $f^{(i)}_t$ can not only effectively comprehend words such as ``walk" and ``kitchen" but also perceive sub-goals such as ``Walk straight ahead across the room."
Therefore, the agent can obtain a clearer global instruction understanding at different scales and perform better in the navigation process.

\subsection{Peak Attention Loss}
\label{sec:method:pal}
PAL is designed to shape the high-level attention score in the training process.
As we use sub-instruction features as attention keys, it is natural to ask how the high-level attention block chooses sub-instructions to focus on.
In our observation, the MLA module brings the ability to form a global perception of the instruction but may cause the attention dispersion problem on sub-instructions sometimes.
That is, the high-level attention attends to several sub-instructions simultaneously, not just one.
Multiple extreme points in the attention score are not our expectation because the agent may become confused about the current navigation progress.
We hope the agent mainly focuses on a single sub-instruction to obtain clear local guidance on the action prediction.
So, we propose a novel loss function, PAL, to guide the model softly concentrate on a single sub-instruction, strengthening the proper local perception of the instruction.

First, the attention score $\alpha_t\in \mathbb{R}^N$ is extracted from the high-level attention, and we find the index of maximum score $k^* = \text{Argmax}_k \alpha_{t,k}$, where $\alpha_{t,k}$ denotes the $k$-th element in the vector $\alpha_t$.
Then, we construct an expected score vector $\beta_t\in \mathbb{R}^N$.
The proposed loss function is calculated by the mean square error between $\beta_t$ and $\alpha_{t}$:
\begin{equation}
\label{eq:peak_loss2}
    z_k = -\frac{(k-k^*)^2}{2\sigma^2}, k=1,\dots,N
\end{equation}
\begin{equation}
\label{eq:peak_loss3}
    \beta_{t,k} = \frac{e^{z_k}}{\sum^N_{j=1} e^{z_j}}
\end{equation}
\begin{equation}
\label{eq:peak_loss4}
    \mathcal{L}_{peak} = \frac{1}{N}\sum^T_{t=1} \sum^N_{k=1} (\alpha_{t,k}-\beta_{t,k})^2
\end{equation}
Elements in $\beta_t$ are a series similar to a Gaussian curve, and $\sigma$ is a hyper-parameter controlling the shape of the curve (called PAL focusing ratio).
In Eq. (\ref{eq:peak_loss2}) and Eq. (\ref{eq:peak_loss3}), we construct the expected distribution by a Gaussian-like curve.
Nevertheless, this curve can be chosen from various curve types with the ``single peak" shape.
We will discuss the effect of different curve types like linear, quadratic, and cubic descent curves in experiments \ref{sec:experiments:ablation_pal}.
Another notable point is that we do not pass the gradient to the $\text{Argmax}$ operator to ensure the differentiability.

\begin{figure}[t]
\centering
\includegraphics[width=0.48\textwidth]{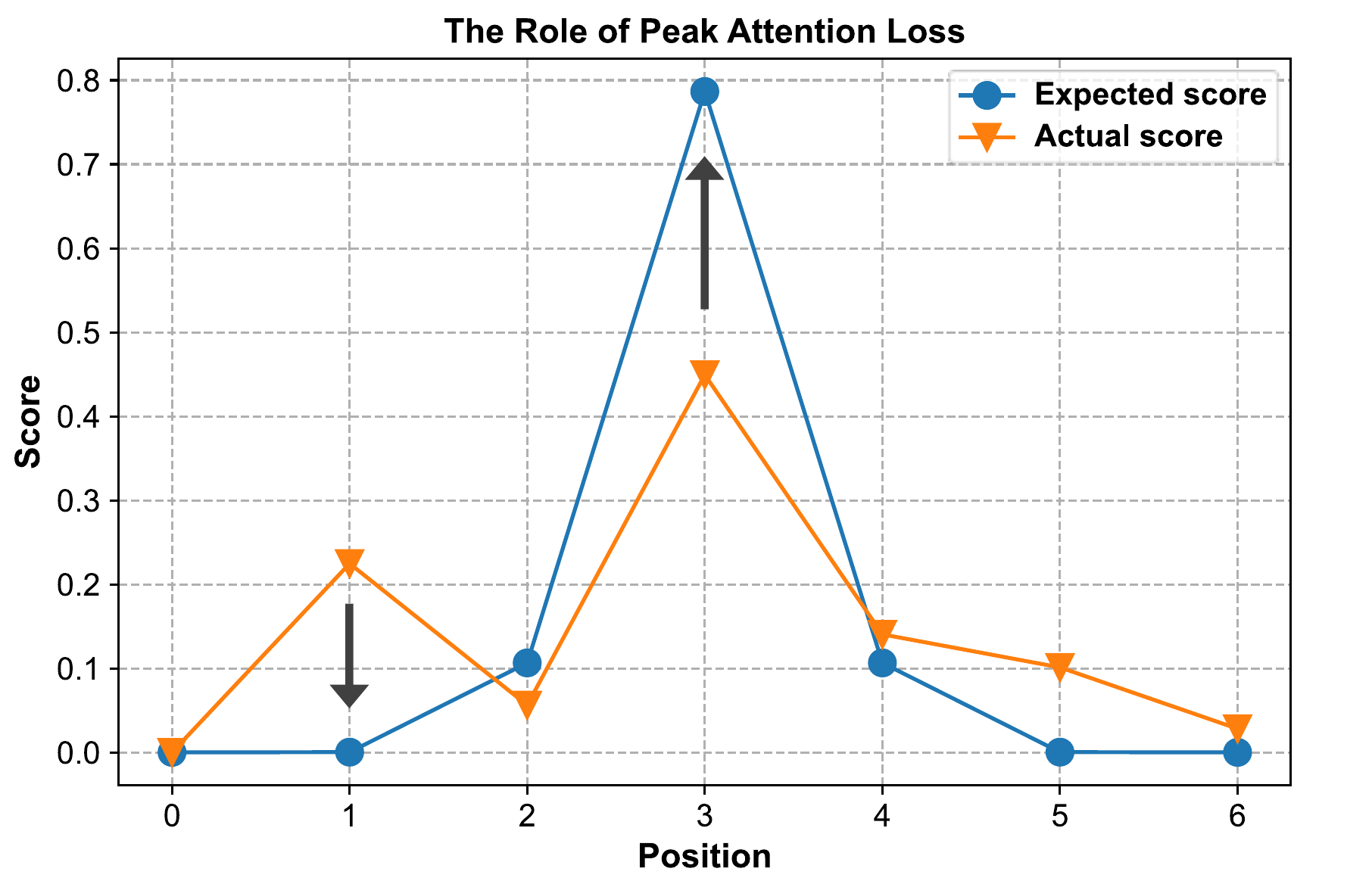}
\caption{The role of PAL. When there are multiple peaks in the actual score (orange line, triangle mark), PAL encourages the global maximum point and weakens other local maximum points, making the actual score closer to the expected score (blue line, round mark).}
\label{fig:peak_loss}
\end{figure}

Fig. \ref{fig:peak_loss} gives an intuitive example of how PAL works.
When the actual attention score $\alpha_t$ has multiple local maximums, PAL forces the distribution to the expected score $\beta_t$ with only one global maximum.
Meanwhile, PAL allows a little information leakage from surroundings around the maximum.
Such leakage ensures a necessary context for the agent to decide whether to shift the attention.
Overall, PAL helps the MLA module's training and benefits the alignment between sub-instruction and the current navigation progress.

\subsection{Action Decoder}
\label{sec:method:action_loss}
The action decoder is a GRU module, which accepts all latent features and output the predicted action:
\begin{equation}
\label{eq:action_decoder}
    h^{(a)}_t = {\rm GRU}_a({\rm Concat}(h^{(h)}_t,h^{(l)}_t,\tilde{a}_{t-1},f^{(i)}_t,f^{(v)}_t);h^{(a)}_{t-1})
\end{equation}
\begin{equation}
\label{eq:action_final1}
    \overline{a}_t = {\rm Softmax}({\rm Linear}(h^{(a)}_t))
\end{equation}
\begin{equation}
\label{eq:action_final2}
    a_t = {\rm Argmax}_k(\overline{a}_{t, k})
\end{equation}

The action distribution $ \overline{a}_t$ will be used for constructing the action loss $\mathcal{L}_{action}=\text{CrossEntropy} (\overline{a}_{t}, \hat{a}_t)$, where $\hat{a}_t$ is the ground truth.
Following Ma et al. \cite{ma2019self} and Zhu et al. \cite{zhu2020vision}, we also employ the progress monitor loss $\mathcal{L}_{progress}$.
Combining these two losses and PAL, we form the total training loss as:
\begin{equation}
\label{eq:total_loss}
    \mathcal{L} = \mathcal{L}_{action}+ \lambda \mathcal{L}_{peak} + \theta \mathcal{L}_{progress}
\end{equation}
$\lambda$ and $\theta$ are parameters balancing different components.
Empirically, we choose $\lambda=0.4$ and $\theta=1.0$ as the best setting after the grid search. 
For $\lambda$, we use a linear-increasing strategy (from $0$ to $0.4$) to control this parameter in the training process to avoid a too strong constraint. 

\section{Experiments}
\label{sec:experiments}

\begin{table*}[htbp]
  \centering
  \caption{Experiment results on validation sets with all metrics. DA and AUG mean DAgger training and EnvDrop augmentation training respectively. Val-Seen contains same scenes as training, while Val-Unseen contains new unseen scenes. Bold font notes the best value of a metric. $\dagger$: Baselines do not use RL training, so we fine-tune official pretrined weights under same conditions.}
  \label{tab:cmp_val}
  \setlength\tabcolsep{5.5pt}
  \begin{tabular}{@{}lc ccc gggggg gggggg @{}} 
    \toprule
      & &&& \multicolumn{6}{c}{\thhead{Val-Seen}} & \multicolumn{6}{c}{\thhead{Val-Unseen}} \\ 
    \cmidrule(lr){6-11}
    \cmidrule(l){12-17}
    \# &\thhead{MODEL}&\thhead{DA}&\thhead{AUG}&\thhead{RL}&\thunit{TL$\downarrow$}&\thunit{NE$\downarrow$}&\thunit{nDTW$\uparrow$}&\thunit{OSR$\uparrow$}&\thunit{SR$\uparrow$}&\thunit{SPL$\uparrow$}&\thunit{TL$\downarrow$}&\thunit{NE$\downarrow$}&\thunit{nDTW$\uparrow$}&\thunit{OSR$\uparrow$}&\thunit{SR$\uparrow$}& \thunit{SPL$\uparrow$} \\
    \midrule
    \rownumber&\multirow{3}{*}{\makecell{$\text{Seq2Seq}^\dagger$\cite{krantz2020beyond}}} & - & -  &-& 8.34 & 8.48 & 0.47 & 0.32 & 0.22 & 0.21 & 8.93 & 9.28 & 0.40 & 0.28 & 0.17 & 0.15 \\
                         \rownumber&& $\checkmark$ & $\checkmark$ &-& 9.37 & 7.02 & 0.54 & 0.46 & 0.33 & 0.31 & 9.32 & 7.77 & 0.47 & 0.37 & 0.25 & 0.22 \\
                        \rownumber&& $\checkmark$ & $\checkmark$ &$\checkmark$ & \textbf{7.30} & 7.55 & 0.51 & 0.38 & 0.30 & 0.28 & \textbf{6.19} & 7.97 & 0.48 & 0.26 & 0.21 & 0.20 \\ 
                         \midrule
\rownumber&\multirow{4}{*}{\makecell{$\text{CMA}^\dagger $\cite{krantz2020beyond}}}     & - & - &-& 8.51 & 8.17 & 0.47 & 0.35 & 0.28 & 0.26 & 7.87 & 8.72 & 0.44 & 0.28 & 0.21 & 0.19 \\
                         \rownumber&& - & $\checkmark$ &-& 8.49 & 8.29 & 0.47 & 0.36 & 0.27 & 0.25 & 7.68 & 8.42 & 0.46 & 0.30 & 0.24 & 0.22 \\
                         \rownumber&& $\checkmark$ & $\checkmark$ &-& 9.06 & 7.21 & - & 0.44 & 0.34 & 0.32 & 8.27 & 7.60 & - & 0.36 & 0.29 & 0.27 \\
                        \rownumber&& $\checkmark$ & $\checkmark$ & $\checkmark$ & 8.22 & 7.63 & 0.51 & 0.34 & 0.25 & 0.23 & 7.70 & 7.77 & 0.49 & 0.32 & 0.25 & 0.23 \\ 
                         \midrule
\rownumber&\multirow{5}{*}{\makecell{MLANet (Ours)}}     &-  &-  &-& 8.20    & 7.56     & 0.51     & 0.37     & 0.30      & 0.28      & 7.64      & 8.32      & 0.46      & 0.29       & 0.23      & 0.21     \\ 
                         \rownumber&&-  &$\checkmark$  &-& 8.49     & 7.10     & 0.55     & 0.44     & 0.35      & 0.33      & 7.51      & 7.98      & 0.48      & 0.30      & 0.24      & 0.23      \\ 
                         \rownumber&&$\checkmark$  &-  &-& 9.70     & 7.10      & 0.52     & 0.46     & 0.33     &    0.30  & 9.53     & 7.67     & 0.48     & 0.39     & 0.28      & 0.26      \\ 
                         \rownumber&&$\checkmark$  &$\checkmark$  &-&  9.07   & 6.04  &    0.58  & \textbf{0.52}      & 0.42     & 0.39     & 8.64     & 7.24  & 0.52     & 0.40     & 0.32     & 0.30    \\ 
                         
\rownumber& &$\checkmark$  &$\checkmark$&$\checkmark$  &  8.10   & \textbf{5.83}  &    \textbf{0.60}  & 0.50      & \textbf{0.44}     & \textbf{0.42}     & 7.21    & \textbf{6.30}  & \textbf{0.58}     & \textbf{0.42}    & \textbf{0.38}     & \textbf{0.35}    \\ 
                         
    \bottomrule
  \end{tabular}
\end{table*}

\subsection{Experiment Setup}
\label{sec:experiments:setup}
\subsubsection{Datasets}
We use VLN-CE dataset with proposed sub-instruction set FSASub to train and evaluate the proposed MLANet model. 
VLN-CE \cite{krantz2020beyond} is the reconstruction of R2R \cite{anderson8578485} dataset in Habitat Simulator \cite{savva2019habitat}, with totally 4475 continuous trajectories and 13425 corresponding instructions.
In the simulator, the agent is attached with an egocentric RGBD camera, with the resolution of $256\times 256$ (RGB image is downsampled to $224\times 224$ to fit encoders) and the horizontal field-of-view of 90\degree.
All samples in VLN-CE are separated into four splits: train, val-seen, val-unseen and test.
Following VLN-CE, the agent can take four types of actions: move forward 0.25m, turn right 15\degree, turn left 15\degree or stop.
However, the original VLN-CE dataset does not contain sub-instructions.
We use the proposed FSA in Section \ref{sec:method} to obtain a dataset containing sub-instructions called ``FSASub".
To prove the effectiveness of FSA, we also reimplement the chunking method in FGR2R \cite{hong2020sub} and use it to generate a dataset called ``FGSub".
Comparison experiments are shown in the following sections.

\subsubsection{Evaluation Metrics}
We use five metrics as previous works \cite{anderson8578485,krantz2020beyond} to evaluate the performance of models:
\begin{itemize}
    \item \textbf{TL}: Trajectory Length, which means the total length of the actual path.
    \item \textbf{NE}: Navigation Error, which means the distance between the stop position and the target position.
    \item \textbf{nDTW}: Normalized Dynamic-Time Warping, which measures the similarity between two time series.
    \item \textbf{OSR}: Oracle Success Rate, the success rate under oracle stopping rule.
    \item \textbf{SR}: Success Rate. When the agent gets within 3 meters of the target, the trajectory is judged as success.
    \item \textbf{SPL}: Success weighted by inverse Path Length is success rate divided by the trajectory length. Following previous works \cite{anderson2018evaluation,anderson8578485,krantz2020beyond}, we use SPL to choose the best model.
\end{itemize}

\subsubsection{Training Details}
We use \texttt{ViT-B/32} version of CLIP as RGB image encoder and sub-instruction encoder.
The dimension of projected RGB features, depth features, word features and sub-instruction features are all 256.
All GRU hidden states and fused features in the model are 512 dimensional.
To alleviate the problem of over-fitting and enhance the model's generalization, we use random dropout on features with ratio 0.25.
The basic model is trained by imitation learning, which means the model tries to mimic the expert actions in the dataset.
We use Adam as the training optimizer, with batch size set to 5 and learning rate set to 2.5e-4. 
In addition, we use 3.2 as the inflection weighting coefficient \cite{wijmans2019embodied} to encourage the agent to make more diverse decisions. 
Because PAL is a strong constrain and may bring noises in early training steps, we use linear increasing strategy for the hyper-parameter $\lambda$. 
We train models for 45 epochs, evaluate models every 5 epochs and choose the best model on the val-unseen split to analyze, consistent with previous work \cite{anderson8578485,krantz2020beyond}.
Besides basic models, we further use Dataset Aggregation (DAgger) \cite{ross2011reduction}, Speaker augmentation \cite{tan2019learning} and Proximal Policy Optimization \cite{schulman2017proximal} to train the best model for comparison with current methods.
The best model is trained with augmentation training (action loss, progress loss and PAL) and DAgger training (action loss and PAL) and fine-tuned with PPO.
We train models on several GPU servers to mitigate the device bias, including GeForce RTX 3090, GeForce RTX 2080 and Tesla T4.
More training details can be found in configuration files in our code repository.

\subsection{Comparison with Baselines}
\label{subsec:baseline}
In Table \ref{tab:cmp_val}, we show the performance of our approach MLANet compared with baselines \cite{krantz2020beyond}.
Seq2Seq\cite{krantz2020beyond} model is a very simple model without dynamic instruction understanding.
CMA\cite{krantz2020beyond} model is equipped with soft attention for dynamic instruction understanding.
For a fair comparison, we also adopt the RNN-based structure used in Seq2Seq and CMA models as our framework. 
The experiments have shown the effectiveness of our proposed method and achieved better performance than the previous models.
Auxiliary methods DA and AUG mean DAgger training and EnvDrop augmentation training, respectively.
RL means fine-tuning the model with a few PPO steps.
Unfortunately, baseline methods do not release their pre-trained weight, so we cannot fine-tune them under the RL setting.

\subsubsection{Without auxiliary training methods}
We find that MLANet is better than Seq2Seq and CMA when trained with only imitation learning.
Compared with Seq2Seq (line 1 vs. 8), MLANet improves SPL from 0.21 to 0.28 (33\% relative) on val-seen and improves SPL from 0.15 to 0.21 (40\% relative) on val-unseen. 
Compared with CMA (line 4 vs. 8), MLANet improves SPL from 0.26 to 0.28 (8\% relative) on val-seen and improves from SPL 0.19 to 0.21 (10\% relative) on val-unseen. 
For other metrics, the MLANet model also achieves improvements.

\subsubsection{With auxiliary training methods}
We also train our MLANet with auxiliary training methods.
The AUG training (line 9 vs. 8) helps MLANet achieve 0.23 SPL on val-unseen, improving 0.02 SPL (10\% relative).
The DA training (line 10 vs. 8) helps MLANet achieve 0.26 SPL on val-unseen, improving 0.05 SPL (24\% relative).
There is an apparent difference between the effects of AUG and DA.
AUG improves a lot of SR on val-seen but lacks generalization ability on val-unseen.
DA improves a lot of SR on val-unseen but performs not as well as AUG on val-seen.
The reason is that AUG training contains many samples in seen scenes that benefit the navigation on val-seen, while DA training encourages exploration and improves generalization performance on unseen environments.
When these two auxiliary methods are applied simultaneously, the MLANet model's performance is significantly improved, achieving 0.39 SPL on val-seen and 0.30 SPL on val-unseen.
After fine-tuned with RL, the performance increases more obviously, achieving 0.35 SPL on val-unseen.

MLANet also outperforms baselines under the same training setting.
First, MLANet (AUG) achieves better SPL than Seq2Seq (DA, AUG) on val-seen and val-unseen (line 9 vs. 2).
Second, MLANet (AUG) achieves better SPL than CMA (AUG) on val-seen and val-unseen (line 9 vs. 5).
Third, MLANet (DA, AUG) achieves better SPL both on val-seen and val-unseen (line 11 vs. 2,6).
Finally, when fine-tuned with reinforcement learning, MLANet (DA, AUG, RL) outperforms than baselines with significant margin (line 12 vs. 3,7).

An surprising phenomenon is that Seq2Seq and CMA both become worse after fine-tuned with reinforcement learning unlike the improvement of MLANet.
Although Seq2Seq (DA, AUG, RL) has the shortest trajectory length, its poor success rate proves that short trajectories are cause by stopping wrongly before reaching the target.
This phenomenon suggest that MLANet has more potential than baselines, which can be released by a stronger training method.

In conclusion, experiment results demonstrate that the proposed model achieves better performance on almost all metrics compared with baselines, indicating the effectiveness of our method.

\subsection{Performance on the Leaderboard}
Following the requirements of the VLN-CE challenge, we submitted the result of our best model to the official test leaderboard.
Table \ref{tab:test} gives the performance compared with baselines and several other methods. 
The MLANet model achieves better performance than CM2 \cite{Georgakis_2022_CVPR}, CMA \cite{krantz2020beyond}, LAW \cite{raychaudhuri2021language}, WPN+DN \cite{krantz2021waypoint} and HPN+DN \cite{krantz2021waypoint} on almost all metrics.
Compared with the CMA baseline, MLANet helps the agent better understand instructions and navigation goals, thus conducting more appropriate actions to complete the VLN task.
HPN+DN \cite{krantz2021waypoint} is a hierarchical decision model, requiring dataset modification and a large amount of reinforcement iterations.
Compared with HPN+DN, MLANet also achieves SPL improvement.
The only non-best metric of MLANet is the navigation error (NE), where MLANet is relatively higher than HPN+DN.
We think the reason is that HPN+DN leverages the graph priority in training process and thus moves closer to the target than MLANet, which do not use discrete graph information.

\begin{table}[htbp]
  \centering
  \caption{Comparison on test leaderboard.}
  \label{tab:test}
  \begin{tabular}{@{}l ggggg @{}}
    \toprule
       & \multicolumn{5}{c}{\thhead{Test}} \\ 
    \cmidrule(l){2-6}
    \thhead{MODEL}&\thunit{TL$\downarrow$}&\thunit{NE$\downarrow$}&\thunit{OSR$\uparrow$}&\thunit{SR$\uparrow$}&\thunit{SPL$\uparrow$} \\
    \midrule
CM2 \cite{Georgakis_2022_CVPR}     & 13.85	& 7.74	& 0.39 &	0.31 &	0.24 \\
CMA \cite{krantz2020beyond}      & 8.85     & 7.91        & 0.36       & 0.28      & 0.25 \\
LAW \cite{raychaudhuri2021language}      & 9.67	& 7.69	& 0.38 &	0.28 &	0.25 \\
WPN+DN \cite{krantz2021waypoint}      & 9.68	& 7.49	& 0.36 &	0.29 &	0.25 \\
HPN+DN \cite{krantz2021waypoint}      & 8.02	& \textbf{6.65}	& 0.37 &	0.32 &	0.30 \\
\midrule
MLANet (Ours)  & \textbf{7.42}      & 6.78        & \textbf{0.39}       & \textbf{0.34}      & \textbf{0.32} \\ 
    \bottomrule
  \end{tabular}
\end{table}

\subsection{Module Ablation Study}
\label{sec:experiments:ablation}
\begin{table*}[htbp]
  \centering
  \caption{Module ablation study.}
  \label{tab:ablation}
  
  \begin{tabular}{@{}l gggggg gggggg @{}}
    \toprule
      & \multicolumn{6}{c}{\thhead{Val-Seen}} & \multicolumn{6}{c}{\thhead{Val-Unseen}} \\ 
    \cmidrule(lr){2-7}
    \cmidrule(l){8-13}
    \thhead{MODEL}&\thunit{TL$\downarrow$}&\thunit{NE$\downarrow$}&\thunit{nDTW$\uparrow$}&\thunit{OSR$\uparrow$}&\thunit{SR$\uparrow$}&\thunit{SPL$\uparrow$}&\thunit{TL$\downarrow$}&\thunit{NE$\downarrow$}&\thunit{nDTW$\uparrow$}&\thunit{OSR$\uparrow$}&\thunit{SR$\uparrow$}&\thunit{SPL$\uparrow$} \\
    \midrule
MLANet & \textbf{8.20}     & \textbf{7.56}     & \textbf{0.51}     & 0.37     & \textbf{0.30}      & \textbf{0.28}      & \textbf{7.64}      & \textbf{8.32}      & \textbf{0.46}      & 0.29       & \textbf{0.23}      & \textbf{0.21} \\ \midrule
\ \ w/o FSA  & 8.62 & 7.93 & 0.48 & 0.37 & 0.28 & 0.26 & 8.20 & 8.32 & 0.45 & \textbf{0.30} & 0.21 & 0.19 \\ 

\ \ w/o MLA  & 8.93 & 8.13 & 0.48 & \textbf{0.39} & 0.28 & 0.26 & 8.47 & 8.49 & 0.43 & 0.30 & 0.21 & 0.19 \\ 
\ \ w/o PAL & 8.67 & 7.62 & 0.50 & 0.39 & 0.29 & 0.27 & 8.02 & 8.77 & 0.43 & 0.28 & 0.20 & 0.18 \\ 
    \bottomrule
  \end{tabular}
\end{table*}

To study the effects of different components, we ablate FSA, MLA and PAL respectively in Table \ref{tab:ablation}.
Auxiliary training is really heavy on both time and storage cost, so we conduct all ablation experiments under only simple imitation learning for time efficiency.

When ablating FSA, all high-level instruction features are masked, so there is no sub-instruction information while keeping other structures unchanged.
The experiments show that the model performs poorly on two validation sets without sub-instructions supplied by FSA.

When ablating MLA, instruction features are averaged and concatenated together, so there is no multi-level attention.
When using a simple average strategy to process instruction features, the SR and SPL decrease because the model can not obtain a clear perception of instructions.
This result further proves that a multi-level understanding of instructions is essential for our model.

When ablating PAL, the balancing parameter $\lambda$ is set to zero.
In our method, PAL is a helpful tool to train the MLA module.
After ablating, the decreased SPL shows that PAL is beneficial for training a better model.

In summary, the model performance drops due to the lack of any part in our multi-level instruction understanding procedure.
Thanks to FSA, MLA and PAL, the full model obtains a nuanced understanding of the instruction and has an excellent ability to reach the correct position in a relatively short path. 

\subsection{Performance of FSA}
\label{sec:experiments:subset}

\begin{figure}[tbh]
\centering
\includegraphics[width=0.48\textwidth]{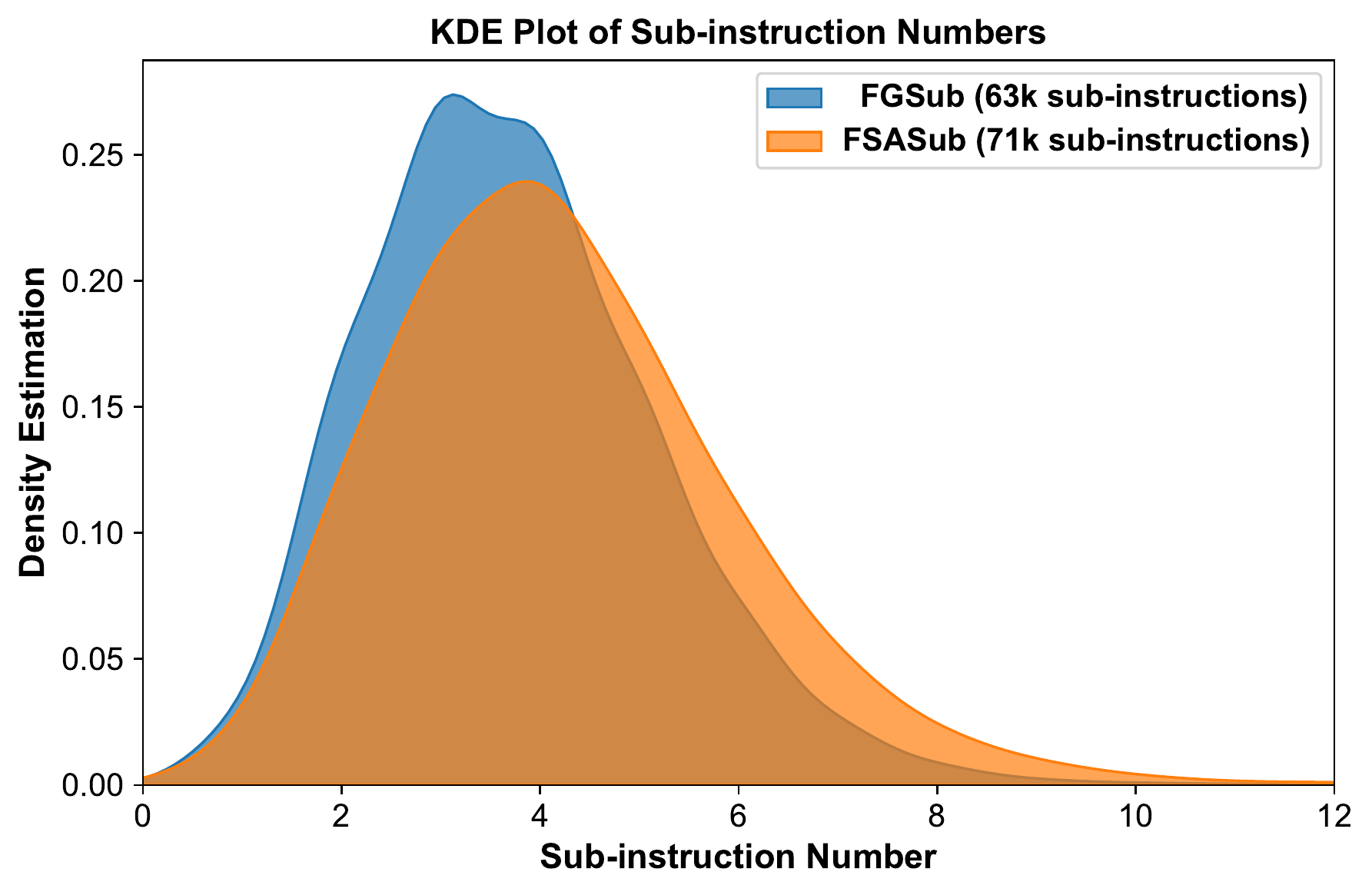}
\caption{The kernel density estimate (KDE) plot of sub-instruction numbers. A higher sub-instruction number means there are more sub-instructions in an instruction.}
\label{fig:cmp_sub_kde}
\end{figure}

FGR2R \cite{hong2020sub} has the closest setting with our method among several works \cite{Hong2020bentity,zhu-etal-2020-babywalk,zhang2021spc-nav}, so we choose it as our comparing subject.
To study the effect of different sub-instruction segmentation methods, we generate two datasets FGSub (from FGR2R \cite{hong2020sub}) and FSASub (from FSA), then use them for training models, respectively. 
In training process, we use hyper parameters $\theta=1.0, \lambda=0.4,  \sigma=0.6$.
Table \ref{tab:sub_set} compares different aspects of these sub-instruction sets, where Time is the time cost of processing the whole dataset; Segment Ratio is the percentage of successfully segmented instructions; Avg Num is the average number of sub-instructions in an instruction.
The result shows that the model trained with FSASub is slightly better than FGSub on both val-seen and val-unseen.
The reason is that FSASub can segment more instructions successfully and supply more sub-instructions for better training.
This advantage is proved by the higher segment ratio and average number of FSASub.
The kernel density estimation (KDE) plot shown in Fig. \ref{fig:cmp_sub_kde} also demonstrates that FSASub has a smoother sub-instruction number distribution than FGSub and thus is more beneficial for training.
Another advantage of FSA is the time efficiency.
Generating FGSub (about 8.8 hours) takes us more than 72 times as long as generating FSASub (only 7.3 minutes).

\begin{table}[htbp]
\centering
\caption{Comparison of two sub-instruction sets. SR and SPL are model performance trained with different sets. }
\label{tab:sub_set}
\setlength\tabcolsep{5pt}
\begin{tabular}{@{} l c c c c c c c @{}}
\toprule
\multirow{2}{*}{\thhead{Name}} & \multicolumn{2}{c}{\thhead{Val-Seen}} & \multicolumn{2}{c}{\thhead{Val-Unseen}}  & \multirow{2}{*}{\thhead{\makecell{Segment \\ Ratio}}} & \multirow{2}{*}{\thhead{\makecell{Avg \\ Num}}} & \multirow{2}{*}{\thhead{\makecell{\color{blue} Time \\ \color{blue} (hours)}}} \\ 
\cmidrule(r){2-3} 
\cmidrule(l){4-5}
                         & \thunit{SR}           & \thunit{SPL}            & \thunit{SR}            & \thunit{SPL}      &     \\ 
                         \midrule
FGSub                    & 0.294             & 0.280  & 0.219              & 0.205    & 97.56\% & 3.72  & {\color{blue} 8.80}        \\ 
FSASub                   & \textbf{0.301}             & \textbf{0.284}  & \textbf{0.228}              & \textbf{0.214}      & \textbf{98.36\%} & \textbf{4.20}  & {\textbf{\color{blue} 0.12}}        \\ 
\bottomrule
\end{tabular}
\end{table}

Here is a qualitative example. Given an instruction, two segmentation methods give different results.
FGR2R method fails to detect the final two sub-instructions connected by ``and" while our method successfully segments them.
That is because our refining rules in FSA can detect the start boundaries of sub-instructions more precisely.

\textbf{FGSub}
\begin{itemize}
  \item {\instfont turn to the right}
  \item {\instfont go past the refrigerator}
  \item {\instfont turn left and walk to the point where you be to the hallway by the entry and dining room area}
\end{itemize}

\textbf{FSASub}
\begin{itemize}
  \item {\instfont Turn to the right}
  \item {\instfont go past the refrigerator}
  \item {\instfont Turn left}
  \item {\instfont and walk to the point where you 're to the hallway by the entry and dining room area}
\end{itemize}

In summary, FSASub is better than FGSub in three aspects.
First, FSASub supplies more effective sub-instructions for model training.
Second, generating FSASub is more time more time-efficient than FGSub.
Finally, models trained with FSASub have better performance than FGSub.
So, we use FSASub as the default sub-instruction set to train our models.

\subsection{Hyper-parameters of MLA}
\label{sec:experiments:ablation_mla}
\begin{figure}[htbp]
\centering
\includegraphics[width=0.48\textwidth]{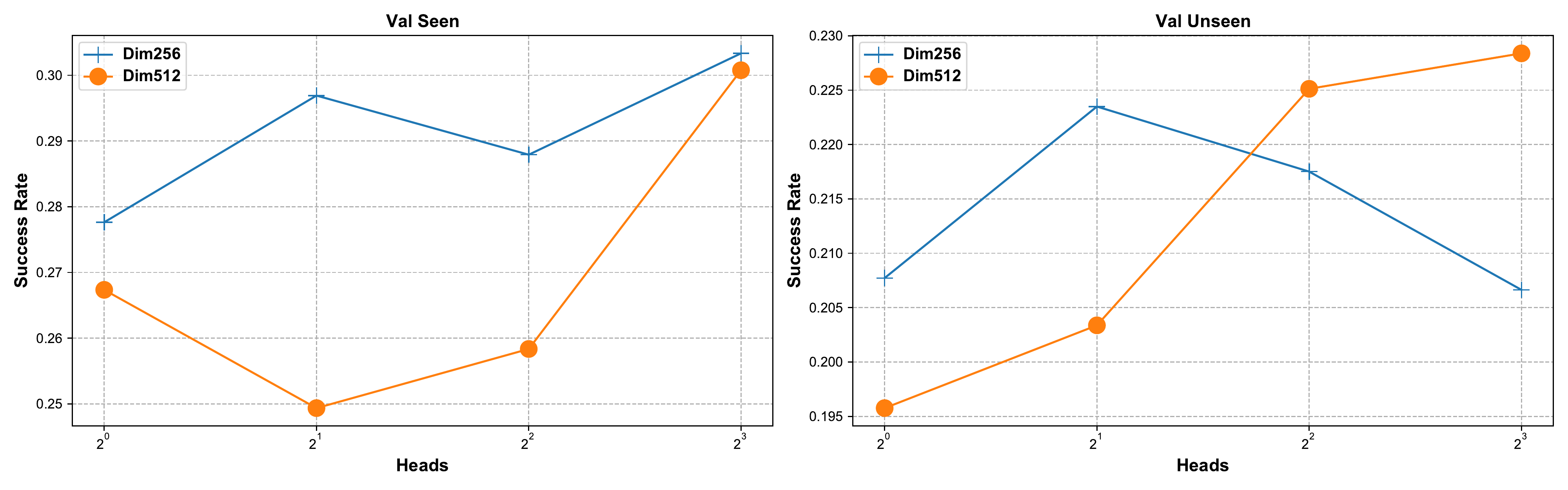}
\caption{Val-seen (left) and val-unseen (right) SR curve of different hyper-parameter settings in MLA. Dim256 means the hidden dimension is 256; Dim512 means the hidden dimension is 512.}
\label{fig:hyper_mla}
\end{figure}
The number of attention heads and hidden dimensions are two key hyper-parameters in MLA.
We study their effects in Fig. \ref{fig:hyper_mla}.
Overall, Dim256 fits well on val-seen but is worse than Dim512 on val-unseen.
Such results indicate that the model with more hidden dimensions generalizes better in the unseen environment due to more substantial learning potential.
As for the number of heads, the result becomes relatively complicated.
Intuitively, more heads would produce better performance because of a more fine-grained understanding of inputs.
This assumption holds on Dim512, where the success rate curve keeps rising as the number of heads increases.
However, when the hidden dimension is set to 256, the success rate on val-unseen reaches a maximum at two heads and begins to decrease.
In a multi-head attention unit, every head has a different focus.
If the total dimension becomes too low and the head number becomes too large, all heads will suffer from inadequate representation ability.
That is why Dim256 cannot keep increasing on val-unseen.
We expect the model to be generalized on unseen environments, so we choose Dim512 with eight heads as the configuration of MLA.

\subsection{Hyper-parameters of PAL}
\label{sec:experiments:ablation_pal}
The curve type and focusing ratio $\sigma$ are two important hyper-parameters for PAL.
We explore their effect in Table \ref{tab:ablation_hyper}.
We compare five curve types: Gaussian, Constant, Linear, Quadratic and Cubic, and we will name models trained under each setting with the corresponding curve type in this part.
Gaussian curve is formulated by Eq. (\ref{eq:peak_loss2},\ref{eq:peak_loss3},\ref{eq:peak_loss4}).
The last four types are polynomials with different degrees, which control the descending trends from the peak to the boundary.
We find that the Gaussian curve achieves the best performance and the Quadratic curve has a competitive performance on val-unseen.
However, Quadratic acts not so well on val-seen, showing less help for a better navigation agent.
For the Gaussian curve we use, the $\sigma$ controls the expected shape of the attention score.
Smaller $\sigma$ shapes the attention score to a sharper peak.
When changing $\sigma$, we find an interesting phenomenon that the model with $\sigma=0.8$ achieves the highest success rate on val-seen.
However, it performs poorly on val-unseen, with a gap of 0.11 SPL between val-seen and val-unseen.
Finally, we decided to use the Gaussian curve with $\sigma=0.6$.
This combination balances hard and soft attention strategies in training, which helps the model achieve good performance.

\begin{table}[htbp]
\centering
\caption{Model performances when using different curve types (TYPE) and PAL focusing ratio ($\sigma$). }
\label{tab:ablation_hyper}
\begin{tabular}{lcccccc}
\toprule
 &&& \multicolumn{2}{c}{\thhead{Val-Seen}} & \multicolumn{2}{c}{\thhead{Val-Unseen}} \\ 
\cmidrule(r){4-5} 
\cmidrule(l){6-7}
                         \thhead{MODEL}&\thhead{TYPE}&\thhead{$\sigma$}& \thunit{SR}           & \thunit{SPL}            & \thunit{SR}            & \thunit{SPL}           \\ 
                         \midrule
\multirow{4}{*}{\makecell{MLANet\\(with PAL)}}& \multirow{4}{*}{Gaussian} & 0.5 &  0.275             & 0.257  & 0.209              & 0.194    \\  
                  && 0.6 &  0.301          & 0.284  & \textbf{0.228}              & \textbf{0.214}    \\ 
                  && 0.8 & \textbf{0.312}            & \textbf{0.295}  & 0.198              & 0.185    \\ 
                  && 1.0 & 0.289            & 0.268  & 0.214              & 0.198    \\   
                         \midrule
\multirow{5}{*}{\makecell{MLANet\\(with PAL)}}& Constant & - &  0.276             & 0.260  & 0.187             & 0.174    \\
& Linear & - &  0.263             & 0.250  & 0.209              & 0.196    \\ 
& Quadratic & - &  0.240             & 0.226  & 0.226              & 0.212    \\  
& Cubic & - &  0.261             & 0.244  & 0.210              & 0.196    \\ 
&Gaussian& 0.6 &  \textbf{0.301}        & \textbf{0.284}  & \textbf{0.228}              & \textbf{0.214}    \\ 
\bottomrule
\end{tabular}
\end{table}

\subsection{Visualization}
\label{sec:experiments:visualization}

\begin{figure}[tbh]
\centering
\includegraphics[width=0.48\textwidth]{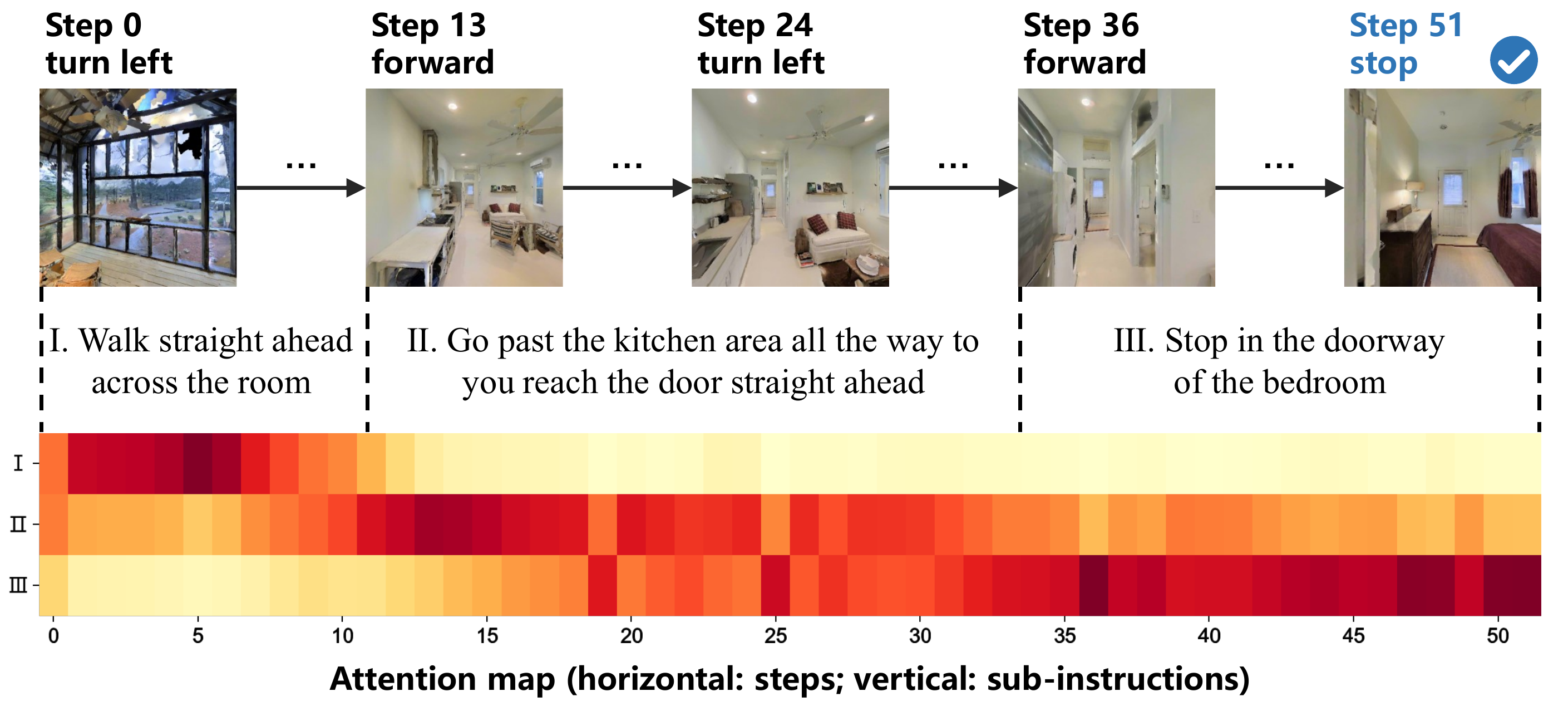}
\caption{A qualitative example of the MLANet model. The upper part shows vision shortcuts and the lower part shows the sub-instruction attention (high-level attention in MLA).}
\label{fig:visualization1}
\end{figure}

Fig. \ref{fig:visualization1} visualizes a path finished by our approach.
The upper part is RGB images for several steps and the lower part is the attention map over time steps.

At the first several steps, the agent concentrates on the sub-instruction ``Walk straight ahead across the room".
However, it is on a balcony and does not see any room.
So, the agent keeps turning left to find a room.
After seeing a room scene, the agent realizes it should cross this room and move forward.
Then, the attention peak gradually shifts to the second sub-instruction ``Go past the kitchen area all the way to you reach the door straight ahead".
At step 13, the agent concentrates on the second sub-instruction and moves forward confidently.
``kitchen" is a keyword for the second part of the path, so the agent turn left at step 24 to check if it is still in the kitchen area.
At step 36, the agent gradually decides to consider the final sub-instruction, ``Stop in the doorway of the bedroom", and move forward to reach the doorway.
Finally, the agent stops at step 51, standing next to a door and seeing the bedroom.
The attention heatmap has a diagonal pattern, which means the model successfully learns to shift the attention over time.
And thanks to PAL, there is only one sub-instruction attended at almost every step.

\begin{figure}[tbh]
\centering
\includegraphics[width=0.48\textwidth]{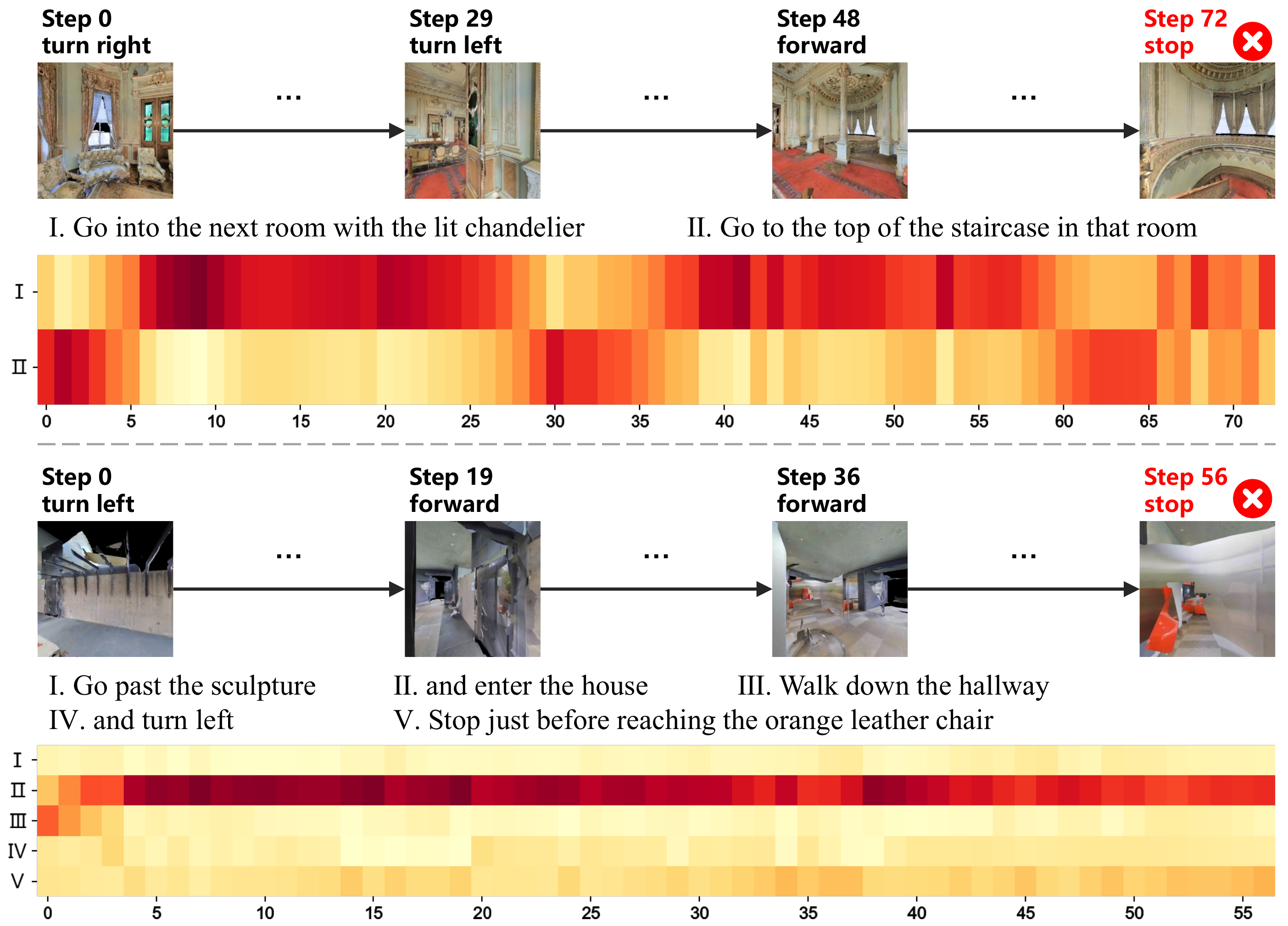}
\caption{Failure cases of sub-instruction attention maps.}
\label{fig:visualization2}
\end{figure}

For a exhaustive analysis of our method, we also two failure cases.
Fig. \ref{fig:visualization2} gives two unsatisfactory attention distributions where the agent fails to reach the target.
One (upper part in Fig. \ref{fig:visualization2}) contains oscillation because the instruction is too simple and contains a rare object, ``lit chandelier". 
The other (lower part in Fig. \ref{fig:visualization2}) only focuses on one sub-instruction all the path.
After examining the corresponding path, we find the path is mainly in a very long hallway, so the model always attends the sub-instruction ``and enter the house."
These failure cases show that our method is sensitive to uncommon language or vision observations, where wrong attention patterns hurt the understanding of the navigation goal and mislead the action prediction.
How to mitigate such phenomenon hints at a future research topic.

\section{Conclusion}
\label{sec:conclusion}
This paper introduces a multi-level instruction understanding procedure and proposes a MLANet for better navigation performance in continuous VLN.
The multi-level instruction understanding procedure contains sub-instruction segmentation, multi-level semantic features fusion and adaptive sub-instruction selection.
We first design a Fast Sub-instruction Algorithm FSA to generate sub-instructions, obtaining a sub-instruction set called ``FSASub''.
FSA is an annotation-free algorithm and achieves significantly higher efficiency than current methods.
We then propose an MLA module, which works for fusing features at the different semantic levels and supplying precise navigation guidance derived from sub-instructions.
Thanks to MLA, the agent obtains a dynamic global perception of the instruction.
Finally, we propose a new loss function PAL, which uses an unsupervised manner to shape the attention score and helps the agent select the current sub-instruction, ensuring a local perception of the current goal.
Experiments show that the model with the MLA module performs better than baselines by 31\%, 30\% and 28\%  SPL improvements on the val-seen, val-unseen and test split, respectively. 
The ablation study and hyper-parameter analysis give more details about our method.

There are still some limitations in this work.
MLANet uses a simple RNN-based framework, which may restrict the sufficient utilization of sub-instructions and observations.
And MLANet is sometimes sensitive to uncommon observations.
Transformer-based frameworks may be our next choice to improve feature representations.
Our future research will focus on introducing a more robust model bottleneck and exploring richer technologies to process sub-instructions. We also plan to experiment with sub-instruction thinking in other VLN or goal-oriented VLN datasets.








\bibliographystyle{IEEEtran}
\bibliography{references}
\end{document}